\documentclass[twoside,11pt]{article}

\usepackage{jmlr2e}

\usepackage{longtable}
\usepackage{multirow}
\usepackage{multicol}
\usepackage{float}
\usepackage{epstopdf}
\usepackage{natbib}
\usepackage{algorithm}
\usepackage{algorithmic}
\usepackage{subfigure}

\usepackage{hyperref}       
\usepackage{url}            
\input{./Definitions.tex}

\newcommand{\m}[1]{\mathbf{#1}}
\newcommand{\ms}[1]{\boldsymbol{#1}}

\usepackage{booktabs}

\newcommand{\Lo}{\mathcal{L}} 

\firstpageno{1}

\begin{document}

\title{Stochastic Sequential Neural Networks  \\ with Structured Inference}

\author{\name Hao Liu \email liuhaosater@gmail.com \\
	   \name Haoli Bai \email haolibai@gmail.com \\
        \addr SMILE Lab \& Yingcai Honors College \\
      University of Electronic Science and Technology of China  \\
       \AND
       \name Lirong He \email lirong\_he@std.uestc.edu.cn \\
       \name Zenglin Xu \email zenglin@gmail.com \\
        \addr SMILE Lab \& Big Data Research Center \\
       School of Computer Science and Engineering \\
       University of Electronic Science and Technology of China
    }

\maketitle

\begin{abstract}
Unsupervised structure learning in high-dimensional time series data has attracted a lot of research interests. For example, segmenting and labelling high dimensional time series can be helpful in behavior understanding and medical diagnosis. 
Recent advances in generative sequential modeling have suggested to combine recurrent neural networks with state space models (e.g., Hidden Markov Models).  This combination can model not only the long term dependency in sequential data, but also the uncertainty included in the hidden states. Inheriting these advantages of stochastic neural sequential models, we propose a structured and stochastic sequential neural network, which models both the long-term dependencies via recurrent neural networks and the uncertainty in the segmentation and labels via discrete random variables.
For accurate and efficient inference, we present a bi-directional inference network by reparamterizing the categorical segmentation and labels with the recent proposed Gumbel-Softmax approximation and resort to the Stochastic Gradient Variational Bayes.
We evaluate the proposed model in a number of tasks, including speech modeling, automatic segmentation and labeling in behavior understanding, and sequential multi-objects recognition. Experimental results have demonstrated that our proposed model can achieve significant improvement over  the state-of-the-art methods.
\end{abstract}


\section{Introduction}
Unsupervised structure learning in high-dimensional sequential data is an important research problem in  a number of applications, such as machine translation, speech recognition,  computational biology, and computational physiology~\cite{sutskever2014sequence,dairecurrent}. For example, in medical diagnosis, learning the segment boundaries and labeling of complicated physical signals is very useful for doctors to understand the underlying behavior or activity types.

Models for sequential data analysis such as recurrent neural networks (RNNs)\cite{rumelhart1988learning} and hidden Markov models (HMMs)\cite{rabiner1989tutorial}  are widely used. Recent literature have investigated approaches of  combining probabilistic generative models and recurrent neural networks for the sake of their complementary strengths in  nonlinear representation learning and effective estimation of parameters~\cite{johnson2016composing,dairecurrent,fraccaro2016sequential}. 
In many tasks, such as segmentation and labeling of natural scenes and physiological signals, the duration lengths and labels of segments are often interpretable and categorical distributed ~\cite{jang2017categorical}. However, most of existing models are designed primarily for continuous situations and do not extend to discrete latent variables ~\cite{johnson2016composing,krishnan2015deep,archer2015black,krishnan2016structured}, probably due to the difficulty of inference for discrete variables in neural networks. For example, the work in \cite{krishnan2015deep} considers combining variational autoencoders~\cite{kingma2013auto} with continuous state-space models, aiming to capture nonlinear dynamics with  control inputs and leading to an RNN-based variational framework.  The work in \cite{johnson2016composing} proposes a state space model with a general emission density. 
When composed with neural networks, state space models are natural to model discrete variables. While discrete variables can be more interpretable and helpful in many applications like medical analysis and behavior prediction, 
they are less considered as switching variables in previous work ~\cite{fox2011bayesian, johnson2016composing}. Although the work in ~\cite{dairecurrent} utilizes discrete variables with informative information for label prediction of segmentation, the inference approach does not explicitly take advantage of structured information to exploit the bi-directional temporal information, and thus may lead to suboptimal performance, as verified in the experiment (See Table~\ref{table_dro} in Section \ref{sec:exp} for more details). 

To address such issues, we propose the Stochastic Sequential Neural Network (SSNN) consisting of a generative network and an inference network.  The generative network(as will shown in Figure~\ref{fig:network}(a)) shares the spirit of Hidden Semi-Markov Model (HSMM) \cite{rabiner1989tutorial}  and Recurrent HSMM~\cite{dairecurrent}, and is composed with a continuous sequence (i.e., hidden states in RNN) as well as two discrete sequences (i.e., segmentation variables and labels in SSM).  The inference network(as will shown in Figure~\ref{fig:network}(b)) can take the advantages of bi-directional temporal information by augmented variables, and efficiently approximate the categorical variables in segmentation and segment labels via the recently proposed Gumbel-Softmax approximation~\cite{jang2017categorical, maddison2016concrete}. Thus, SSNN can model the complex and long-range dependencies in sequential data, but also maintain the structure learning ability of SSMs with efficient inference. 

In order to evaluate the performance of the proposed model, we compare our proposed model with the state-of-the-art neural models in a number of tasks, including automatic segmentation and labeling on datasets of speech modeling, and behavior modeling, medical diagnosis, and multi-object recognition. Experimental results in terms of both model fitting and  labeling of learned segments have demonstrated the promising performance of the proposed model.


\section{Preliminaries}
\label{sec:prelim}

\begin{figure}[!htp]
	\begin{centering}
		\includegraphics[width = 0.89 \linewidth]{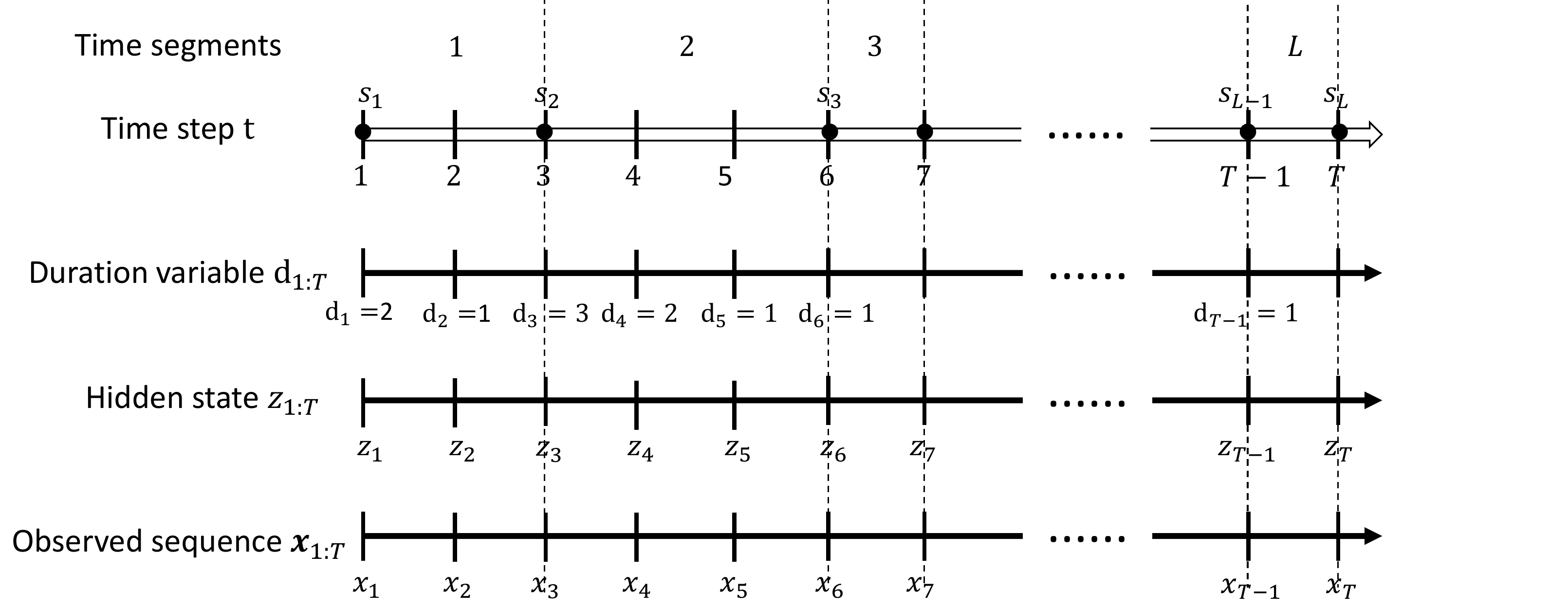}
		\begin{small}
			\caption{
				\small{A visualization of the observed sequence $\m x_{1:T}$ with the corresponding time segments, hidden states $z_{1:T}$ and duration variables $d_{1:T}$. In HMM, there is no duration variable $d_t$ and segments are pre-determined.}
			}
		
			\label{fig:notation}
		\end{small}
	\end{centering}
\end{figure}

In this section, we present background related to generative sequential models. Specifically, we first introduce RNNs, followed by HMMs and HSMMs. 


Recurrent neural networks (RNNs),  
as a wide range of sequential models for time series, have been applied in a number of applications \cite{sutskever2014sequence}. Here we introduce basic properties and notations of RNNs. 
Consider a sequence of temporal sequences of vectors $ \m x_{1:T} = [\m x_1, \m x_2,...,\m x_T]$ that depends on the inputs $\m{u}_{1:T} = [\m u_1, \m u_2,..., \m u_T]$, where $\m x_t \in \mathbb{R}^m$ is the observation and $\m u_t\in \mathbb{R}^{n}$ is the input at the time step $t$, and $T$ is the maximum time steps.  RNN further introduces hidden states $\m{h}_{1:T}=[\m h_1, \m h_2,..., \m h_T]$, where $\m h_t\in \mathbb{R}^{h}$ encodes the information before the time step $t$, and is determined by $\m h_t = f_\theta(\m h_{t-1},\m u_t)$, where $f_\theta(\cdot )$ is a nonlinear function parameterized by a neural network. 

State space models such as hidden Markov models (HMMs) and hidden Semi-Markov models (HSMMs) are also widely-used methods in sequential learning \cite{dairecurrent,chiappa2014explicit,dewar2012inference}. 
In HMM, given an observed sequence $\m{x}_{1:T}$, each $\m x_t$ is generated based on the hidden state $z_t\in\{1, 2, ..., K\}$, and $p_\theta(\m x_t|z_t)$ is the emission probability.
We set $p_\theta(z_1)$ as the distribution of the initial state, and $p_\theta(z_t|z_{t-1})$ is the transition probability. 
We use $z_{1:T} = [z_1, z_2, ..., z_T]$. Here $\theta$ includes all parameters necessary for these distributions.
HSMM is a famous extension of HMM. Aside from the hidden state $z_t$, HSMM further introduces time duration variables $d_t \in \{1,2,..,M\}$, where   $M$ is the maximum duration value for each $\m x_t$. We set $d_{1:T} = [d_1, d_2,...,d_T]$. HSMM splits the sequence into $L$ segments, allowing the flexibility to find the best segment representation. We set $\m{s}_{1:L}=[s_1, s_2,..,s_L]$ as the beginning of the segments. A difference from HMM is that for segment $i$, the latent state $z_{s_i:s_i+d_{s_i}-1}$ is fixed in HMM. An illustration is given in Figure \ref{fig:notation}. 

There are many variants of HSMMs such as the Hierarchical Dirichlet-Process HSMM (HDP-HSMM) \cite{johnson2013bayesian} and subHSMM \cite{johnson2014stochastic}.
The subHSMM and HDP-HSMM extend their HMM counterparts by allowing explicit modeling of state duration lengths with arbitrary distributions. While there are various types of HMM, the inference methods are mostly inefficient. 

Although HMMs and HSMMs can explicitly model uncertainty in the latent space and learn a interpretable representation through $d_t$ and $z_t$, they are not good at capturing the long-range temporal dependencies when compared with RNNs. 

\section{Model}
\label{sec:model}
In this section, we present our stochastic sequential neural network model. The notations and settings are generally consistent with HSMM Section 2, as also illustrated in Figure~\ref{fig:notation}. For the simplicity of explanation, we present our model on a single sequence. It is straightforward to apply the model multiple sequences.

\subsection{Generative Model} 
\label{generative}

\begin{figure*}[h]
	\centering
	\subfigure[Generative network]{
		\includegraphics[width = 0.40 \linewidth]{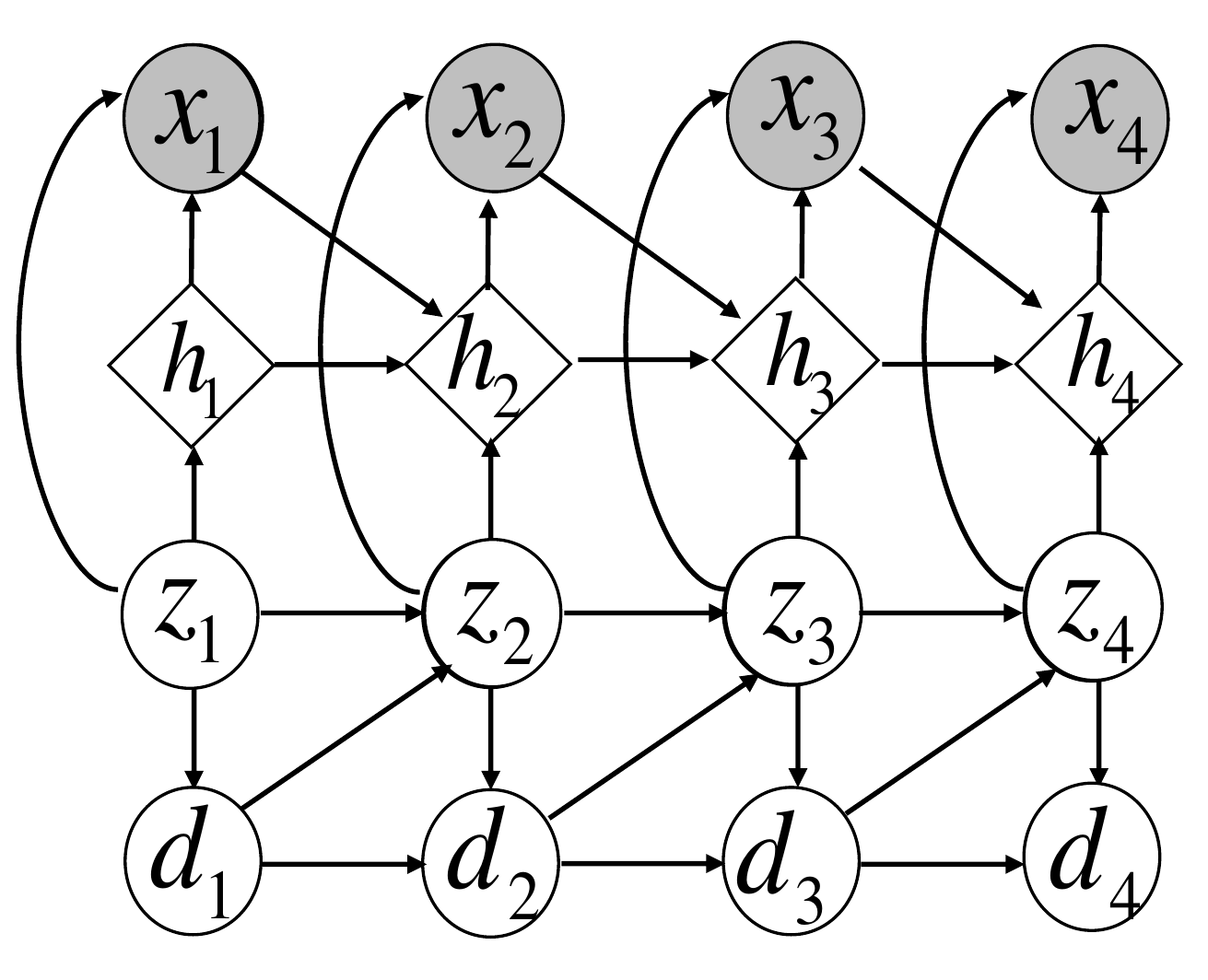}\label{fig:generative}} \hspace{12ex}
	\subfigure[Inference network]{
		\includegraphics[width = 0.38\linewidth]{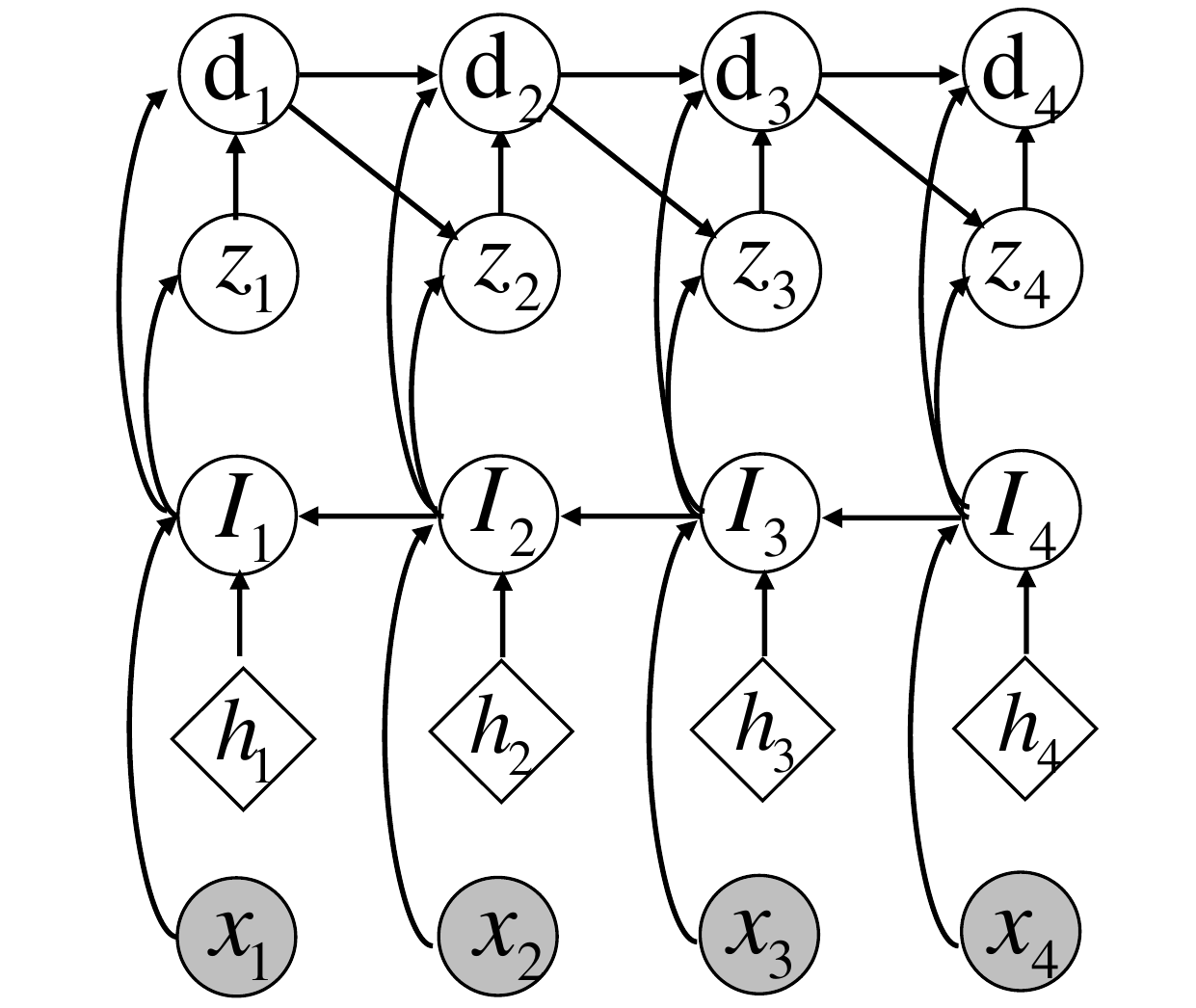}\label{fig:inference}}
	\caption{
		\small{The generative network and inference network of SSNN. The generative network can be viewed as a HSMM with recurrent structure on hidden states. The inference network is designed by backwards recurrent function through $\Ical_t$ for each segment. Diamond units are deterministic variables, while circles are random variables.}
	}
	\label{fig:network}
\end{figure*}

In order to model the long-range temporal dependencies and the uncertainty in segmentation and labeling of time series, we aim to take advantages from RNN and HSMM, and learn categorical information and representation information from the observed data recurrently. As illustrated in Figure~\ref{fig:generative}, we  design an Stochastic Sequential Neural Network (SSNN) with one sequence of continuous latent variables modeling the recurrent hidden states, and two sequences of discrete variables denoting the segment duration and labels, respectively. The joint probability can be factorized as:
\vspace{-1ex}
\begin{align}
	\label{eq-logli}
	p_\theta(\m x_{1:T}, z_{1:T}, d_{1:T}) = p_\theta(\m x_{1:T}|z_{1:T}, d_{1:T})\cdot p_\theta(z_1)p_\theta(d_1|z_1) \prod_{t=2}^{T}p_\theta(z_t|z_{t-1},d_{t-1}) p_\theta(d_t|z_t, d_{t-1}).
\end{align}

To learn more interpretative latent labels, we follow the design in HSMM to set $z_t$ and $d_t$ as categorical random variables,
The distribution of $z_t$ and $d_t$ is
\begin{equation}
p_\theta(z_t|z_{t-1}, d_{t-1})=\left\{
\begin{aligned}
\mathbb{I}(z_t=z_{t-1}) \hspace{2ex} \text{if}\hspace{1ex} d_{t-1} > 1 \\
p_\theta(z_t|z_{t-1}) \hspace{5ex} \text{otherwise}
\end{aligned}\hspace{3.5ex},
\right.
\end{equation}
\begin{equation}
p_\theta(d_t|z_t, d_{t-1})=\left\{
\begin{aligned}
\mathbb{I}(d_t=d_{t-1}-1) \hspace{2ex} \text{if}\hspace{1ex} d_{t-1} > 1 \\
p_\theta(d_t|z_t) \hspace{11ex} \text{otherwise}
\end{aligned}
\hspace{2ex},
\right.
\end{equation}
where $\mathbb{I}(x)$ is the indicator function (whose value equals $1$ if $x$ is True, and otherwise $0$). The transition probability $p_{\theta}(z_t|z_{t-1})$ and  $p_{\theta}(d_t|z_t)$, in implementation, can be achieved by learning a transition matrix.

The joint emission probability $p_\theta(\m x_{1:T}|z_{1:T}, d_{1:T})$ can be further factorized into multiple segments. 
Specifically, for the $i$-th segment $\m x_{s_i:s_i+d_{s_i}-1}$ starting from $s_i$, the corresponding generative distribution is
\vspace{-2ex}
\begin{equation}
p_\theta(\m x_{s_i:s_i+d_{s_i}-1} | z_{s_i}, d_{s_i} ) 
= \prod_{t=s_i}^{s_i+d_{s_i}-1}p_\theta(\m x_t|\m x_{s_i:t-1}, z_{s_i}).
= \prod_{t=s_i}^{s_i+d_{s_i}-1}p_\theta(\m x_t|\m h_t,z_{s_i}),
\end{equation}

where $\m h_t$ is the latent deterministic variable in RNN. As mentioned earlier, $\m h_t$ can better model the complex dependency among segments, and capture past information of the observed sequence $\m x_{t-1}$ as well as the previous state $\m h_{t-1}$. We design $\m h_t = \sigma(\m W_x^{(z_{s_i})}\m x_{t-1} + \m W_h^{(z_{s_i})}\m h_{t-1} + \m b_h^{(z_{s_i})})$, where $\sigma()$ is a tanh activation function, $\m W_x \in \mathbb{R}^{K\times h \times m}$ and $\m W_h\in \mathbb{R}^{K\times h \times h}$ are weight parameter, and $\m b_h\in \mathbb{R}^{K\times h}$ is the bias term. $\m W_x^{(z_{s_i})}\in \mathbb{R}^{h\times m}$ is the $z_{s_i}$-th slice of $\m W_x$, and it is similar for $\m W_h^{(z_{s_i})}$ and $\m b_h^{(z_{s_i})}$.

Finally, the distribution of $\m x_t$ given $\m h_t$ and $z_{s_i}$ is designed by a Normal distribution,
\begin{equation}
p_\theta(\m x_t|\m h_t,z_{s_i})= \mathcal{N}(x; \ms\mu, \ms\sigma^2),
\end{equation}
where the mean satisfies 
$\ms\mu = \m W_{\mu}^{(z_{s_i})}\m h_t+b_{\mu}^{(z_{s_i})}$, and the covariance is a diagonal matrix with its log diagonal elements $\log\ms\sigma^2 = \m W_\sigma^{(z_{s_i})}\m h_t + \m b_{\sigma}^{(z_{s_i})}
$. We use $\theta$ to include all the parameters in the generative model.

\subsection{Structured Inference} 
\label{sec:inference}

We are interested in maximizing the marginal log-likelihood $\log p(\m x)$, however, this is usually intractable since the complicated posterior distributions cannot be integrated out generally. Recent methods in Bayesian learning, such as the score function (or REINFORCE) \cite{archer2015black} and the Stochastic Gradient Variational Bayes (SGVB) \cite{kingma2013auto}, are common black-box methods that lead to tractable solutions. We resort to the SGVB since it could efficiently learn the approximation with relatively low variances \cite{kingma2013auto}, while
the score function suffers from high variances and heavy computational costs. 

We now focus on maximizing the evidence lower bound also known as $ELBO$,
\begin{equation}
\log p_{\theta}(\m x) \geq \Lo(\m x_{1:T};\theta,\phi) = E_{q_\phi(z_{1:T},d_{1:T}|\m x_{1:T})} [ \log p_{\theta}(\xb_{1:T} , z_{1:T} , d_{1:T} ) -\log q_\phi(z_{1:T}, d_{1:T}|\xb_{1:T})],
\label{eq-elbo}
\end{equation}
where $q_\phi(\cdot)$ denotes the approximate posterior distribution, and $\theta$ and $\phi$ denote parameters for their corresponding distributions, repsectively.

\subsubsection{Bi-directional Inference}

In order to find a more informative approximation to the posterior, we augment both random variables $d_t$, $z_t$ with bi-directional information in the inference network. 
Such attempts have been explored in many previous work \citep{krishnan2016structured,khan2017conjugate,krishnan2015deep}, however they mainly focus on continuous variables, and little attention is paid to the discrete variable. Specifically, we first learn a bi-directional deterministic variable $\hat{\m h_t} = \text{BiRNN}(x_{1:t}, x_{t:T})$ , where \textrm{BiRNN} is a bi-directional RNN with each unit implemented as an LSTM \cite{hochreiter1997long}. Similar to \cite{fraccaro2016sequential}, we further use a backward recurrent function $I_t = g_{\phi_I}(I_{t+1}, [\m x_t,\hat{\m h_t}])$ to explicitly capture forward and backward information in the sequence via $\hat{\m h_t}$, where $[\m x_t,\hat{\m h_t}]$ is the concatenation of $\m x_t$ and $\hat{\m h_t}$.

The posterior approximation can be factorized as
\begin{align}
	\label{eq:app}
	& q_\phi(z_{1:T},d_{1:T}|\m x_{1:T})  = q_\phi(z_1|I_1)q_\phi(d_1|z_1,I_1)\prod_{t=2}^{T}  q_\phi(z_t|d_{t-1},I_t)q_\phi(d_t|d_{t-1},z_t,I_t),
\end{align}
and the graphical model for the inference network is shown in Figure.\ref{fig:inference}.
We use $\phi$ to denotes all parameters in inference network.

We design the posterior distributions of $d_t$ and $z_t$ to be categorical distributions, respectively, as follows:
\begin{align}
	\label{eq:q_z}
	q(z_t|d_{t-1},I_t;\phi) = \mathrm{Cat}(\text{softmax}(\m W_z^TI_t)), \\ 
	\label{eq:q_d}
	q(d_t|d_{t-1},z_t,I_t;\phi) =\mathrm{Cat}(\text{softmax}(\m W_d^T I_t)), 
\end{align}
where $\mathrm{Cat}$ denotes the categorical distribution. Since the posterior distributions of $z_t$ and $d_t$ are conditioned on $I_t$, they depend on both the forward sequences (i.e., $h_{t:T}$ and $x_{t:T}$) and the backward sequences (i.e., $h_{1:t-1}$ and $x_{1:t-1}$), leading to a more informative approximation. However, the reparameterization tricks  and their extensions \citep{chung2016hierarchical} are not directly applicable due to the discrete random variables, i.e.,  $d_t$ and $z_t$ in our model. Thus we  turn to the recently proposed Gumbel-Softmax reparameterization trick \citep{jang2017categorical,maddison2016concrete}, as shown in the following.

\subsubsection{Gumbel-Softmax Reparameterization}

The Gumbel-Softmax reparameterization proposes an
alternative to the back propagation through discrete random variables via the Gumbel-Softmax distribution, and circumvents the non-differentiable categorical distribution.

To use the Gumbel-Softmax reparameterization, we first map the discrete pair $(z_t, d_t)$ to a $N$-dimensional vector $\ms\gamma(t)$, and $\ms\gamma(t)\sim \mathrm{Cat}(\ms\pi(t))$, where $\ms\pi(t)$ is a $N$-dimensional vector on the simplex and $N=K\times D$. Then we use $\m y(t)\in \mathbb{R}^{N}$ to represent the Gumbel-Softmax distributed variable:
\begin{equation}
\label{eq-gum}
y_i(t) =  \frac{\text{exp}((\log(\pi_i(t))+g_i)/\tau)}{\sum_{j=1}^k \text{exp}((\log(\pi_j(t))+g_j)/\tau)} \qquad \text{for } i=1, ..., N,
\end{equation}
where $g_i\sim \mathrm{Gumbel}(0, 1)$, and $\tau$ is the temperature that will be elaborated in the experiment. Via the Gumbel Softmax transformation, we set  $\m y(t)\sim \textrm{Concrete}(\ms\pi(t),\tau)$ according to \citep{maddison2016concrete}.

Now we can sample $\m y(t)$ from the Gumbel-Softmax posterior in replacement of the categorically distributed $\ms\gamma(t)$, and use the back-propagation gradient with the ADAM \cite{kingma2014adam} optimizer to learn parameters $\theta$ and $\phi$.

For simplicity, we denote $F(z,d) = \log p_{\theta}(\xb_{1:T} , z_{1:T} , d_{1:T}) -\log q(z_{1:T}, d_{1:T}|\xb_{1:T})$, and furthermore, $\tilde{F}(y,g)$ is the corresponding approximation term of $F(z,d)$ after the Gumbel-Softmax trick. The Gumbel-Softmax approximation of $\Lo(\m x_{1:T};\theta,\phi)$ is:
\begin{align}
	\Lo(\m x_{1:T};\theta,\phi) \approx E_{y \sim \mathrm{Concrete}(\ms\pi(t),\tau)} [ \tilde{F}(y, g) ]  = E_{g \sim \prod_N \mathrm{Gumbel}(0, 1)}[\tilde{F}(y,g)]. 
\end{align}
Hence the derivatives  of the approximated $ELBO$ w.r.t. the inference parameters $\phi$ can be approximated by the SGVB estimator:
\begin{align}
	\label{eq-mc}
	\frac{\partial}{\partial \phi} E_{g \sim \prod_N \mathrm{Gumbel}(0, 1)}[\tilde{F}(y,g) ]
	& = E_{ g \sim \prod_N \mathrm{Gumbel}(0, 1) }\left[\frac{\partial}{\partial \phi} \left(\tilde{F}(y,g) \right)\right] \nonumber \\
	&\approx \frac{1}{B} \sum_{b = 1}^B \frac{\partial}{\partial \phi} \left( \tilde{F}(y^b,g^b)  \right), \hspace{2ex} g^b := (g_1^b, \dotsc, g_N^b),
\end{align}
where $y^b, g^b$ is the batch samples and $B$ is the number of batches. The derivative w.r.t the generative parameters $\theta$ does not require the Gumbel-Softmax approximation, and can be directly estimated by the Monte Carlo estimator
\begin{equation}
\label{eq-theta}
\frac{\partial}{\partial\theta} E_{q_\phi(z_{1:T}, d_{1:T})}[F(z,d) ]\approx
\frac{1}{B} \sum_{b = 1}^B \frac{\partial}{\partial \theta} \left( F(z^b,d^b)  \right). \hspace{2ex}
\end{equation}

Finally, we summarize the inference algorithm in Algorithm~\ref{alg1}.

\begin{algorithm}[h]
	\caption{\small \textbf{Strucutured Inference Algorithm for SSNN}}
	\begin{algorithmic} 
		\STATE \textbf{inputs}: Observed sequences $\{\m x^{(n)}\}_{n=1}^N$
		\STATE \qquad\quad\hspace{0.6ex} Randomly initialized $\phi^{(0)}$ and $\theta^{(0)}$;
		\STATE \qquad\quad\hspace{0.6ex}
		Inference Model: $ q_{\phi}(z_{1:T},d_{1:T}|\m x_{1:T})$;
		\STATE \qquad\quad\hspace{0.6ex} Generative Model: $p_{\theta}(\m x_{1:T}, z_{1:T}, d_{1:T})$;
		\STATE \textbf{outputs}:Model parameters $\theta$ and $\phi$;
		\FOR {$i=1$ to $Iter$}
		\STATE 1. Sample sequences $\{x^{(n)} \}_{n=1}^M $ uniformly from dataset with a mini-batch size $B$.
		\STATE 2. Estimate and sample forward parameters using Eq.(\ref{eq-logli}).
		\STATE 3. Evaluate the \emph{ELBO} using Eq. (\ref{eq-elbo}).
		\STATE 4. Estimate the Monte Carlo approximation to $\nabla_{\theta}L$ using Eq. (\ref{eq-theta})
		\STATE 5. Estimate  the SGVB  approximation to $\nabla_{\phi}L$ using Eq.(\ref{eq-mc}) with the
		Gumbel-Softmax approximation in Eq.(\ref{eq-gum}); 
		\STATE 6. Update $\theta^{(i)}$, $\phi^{(i)}$ using the ADAM.
		\ENDFOR 
	\end{algorithmic}
	\label{alg1}
\end{algorithm}

\section{Related Work}

In this section, we review research work on generative sequential data modeling in terms of state space models and recurrent neural networks. In the following, we review some recent work on sequential latent variable model.

In terms of combining both and SSM and RNNs, the papers mostly close to our paper include \cite{johnson2016composing,krishnan2015deep,krishnan2016structured,archer2015black,fraccaro2016sequential}. In detail, \cite{krishnan2015deep} combines variational auto-encoders with continuous state-space models, emphasizing the relationship to linear dynamical systems. 
\cite{krishnan2016structured} lets inference network conditioned on both future and past hidden variables, which extends Deep Kalman Filtering.
\cite{archer2015black} uses a structured Gaussian variational family to solve the problem of variational inference in general continuous state space models without considering parameter learning. 
And \cite{johnson2016composing} can employ general emission density for structured inference. \cite{fraccaro2016sequential} extends state space models by combining recurrent neural networks with stochastic latent variables. 
Different from the above methods that require the hidden states of SSM  be  continuous, our paper utilizes discrete latent variables in the SSM part for better interpretablity, especially in applications of segmentation and labeling of high-dimensional time series.

In parallel, some research also works on variational inference with discrete latent variables recently. \cite{bayer2014learning} enhances recurrent neural networks with stochastic latent variables which they call stochastic neural network. For stochastic neural network the most applicable approach is the score function or REINFORCE approach, however it suffers from high variance. \cite{mnih2016variational} proposes a gradient estimator for multi-sample objectives that use the mean of other samples to construct a baseline for each sample to decrease variance. \cite{gu2015muprop} also models the baseline as a first-order Taylor expansion and overcomes back propagation through discrete sampling with a mean-field approximation, so it becomes practical to compute the baseline and derive the relevant gradients. \cite{gregor2013deep} uses the first-order Taylor approximation as a baseline to reduce variances. In Discrete VAE \cite{rolfe2016discrete}, the sampling is autoregressive through each binary unit, which allows every discrete choice to be marginalized out in a tractable manner. \cite{dairecurrent} proposes to overcome the difficulty of learning discrete variables by optimizing their distribution instead of directly learning discrete variables.

In the aspect of optimization,
\cite{khan2015kullback,khan2016faster} split the variational inference objective into a term to be linearized and a tractable concave term, which makes the resulting gradient easily to compute.
\cite{knowles2011non} proposes natural gradient descent with respect to natural parameters on each of the variational factors in turn. In \cite{dai2017stochastic}, the discrete optimization is replaced by the maximization over the negative Helmholtz free energy.
In contrast to linearizing intractable terms around the current iteration as used in the above approaches, we handle intractable terms via recognition networks and amortized inference(with the aid of Gumbel-Softmax reparameterization \cite{jang2017categorical,maddison2016concrete}) in this paper. That is, we use parametric function approximators to learn conditional evidence in a conjugate form.

\section{Experiment}
\label{sec:exp}
In this section, we evaluate SSNN on several datasets across multiple scenarios. Specifically, we first evaluate its performance of finding complex structures
and estimating data likelihood on a synthetic dataset and two speech datasets (TIMIT \& Blizard).
Then we test SSNN with learning segmentations and latent labels on
Human activity \cite{reyes2016transition} dataset, Drosophila dataset \cite{kain2013leg} and PhysioNet \cite{springer2016logistic} Challenge dataset, and compare the results with HSMM and its variants. 
Finally we provide an additional challenging test on the multi-object recognition problem using the generated multi-MNIST dataset.

All models in the experiment use the Adam \cite{kingma2014adam} optimizer. Temperatures of Gumbel-Softmax were fixed throughout training. We implement the proposed model based on  Theano~\cite{al2016theano} and Block \& Fuel~\cite{van2015blocks}.
\subsection{Synthetic Experiment} 
To validate that our method is able to model high dimensional data with complex dependency, we simulated a complex dynamic torque-controlled pendulum governed by a differential equation to generate non-Markovian observations from a dynamical system:
$ml^2 \frac{d^2 \phi (t) }{dt^2} = -\mu \frac{d\phi(t)}{dt} + mgl \sin \phi(t) + u(t)$.
For fair comparison with \cite{karl2016deep}, we set  $m = l = 1$, $\mu = 0.5$, and $g = 9.81$. We convert the generated ground-truth angles to image observations.  The system can be fully described by angle and angular velocity.

We compare our method with Deep Variational Bayes Filter(DVBF-LL) \cite{karl2016deep}  and Deep Kalman Filters(DKF) \cite{krishnan2015deep}. 
The ordinary least square regression results are shown in Table ~\ref{exp:pendulum}. Our method is clearly better than DVBF-LL and DKF in predicting $\sin \phi$, $\cos \phi$ and $\frac{d \phi}{dt}$. SSNN achieves a higher goodness-of-fit than other methods. The results indicate that generative model and inference network in SSNN are capable of capturing complex sequence dependency. 

\begin{table*}[htbp]
	\begin{center}
		\begin{small}
			\begin{sc}
				\begin{tabular}{llllllll}
					\hline
					&     & \hspace{6ex} DVBF-LL        &       & \multicolumn{2}{l}{\hspace{8ex} DKF} & \multicolumn{2}{l}{\upshape Our method (SSNN)} \\
					&     & \upshape log-likelihood & $R^2$    & \upshape log-likelihood  & $R^2$     & \upshape log-likelihood     & $R^2$         \\
					\hline
					\upshape Measured   & $\sin \phi$ & 3990.8         & 0.961 & 1737.6          & 0.929 & 4424.6            & 0.975     \\
					\upshape groundtruth & $\cos \phi$ & 7231.1         & 0.982 & 6614.2          & 0.979 & 8125.3             & 0.997     \\
					\upshape variables    & $\frac{d\phi}{dt}$ & -11139         & 0.916 & -20289          & 0.035 & -9620             & 0.941 \\
					\hline
				\end{tabular}
			\end{sc}
		\end{small}
	\end{center}
	\begin{small}
		\caption{
			\small{
				The results measured on the log-likelihood  and  the goodness-of-fit (denoted by $R^2$) given by three methods on the prediction of all latent states on respective dependent variables in pendulum dynamics. For both measures, the higher the better.
			}
			\label{exp:pendulum}
		}
	\end{small}
\end{table*}
\vskip -1.3in

\subsection{Speech Modeling}

We also test SSNN on the modeling of speech data, i.e., Blizzard and TIMIT datasets \cite{prahallad2013blizzard}. Blizzard records the English speech with 300 hours by a female speaker. TIMIT is a dataset with 6300 English sentences read by 630 speakers. For the TIMIT and Blizzard dataset, the sampling frequency is 16KHz and the raw audio signal is normalized using the global mean and standard deviation of the training set.  Speech modeling on these two datasets has shown to be challenging since there's no good representation of the latent states \cite{chung2015recurrent, fabius2014variational, gu2015neural, gan2015deep, sutskever2014sequence}.

The data preprocessing and the performance measures are identical to those reported in \cite{chung2015recurrent, fraccaro2016sequential}, i.e. we report the average log-likelihood for half-second sequences on Blizzard, and report the average log-likelihood per sequence for the test set sequences on TIMIT. 
For the raw audio datasets, we use a fully factorized Gaussian output distribution.

In the experiment, We split the raw audio signals in the chunks of 2 seconds. The waveforms are divided into non-overlapping vectors with size 200.
For Blizzard we split the data using 90$\%$ for training, 5$\%$ for validation and 5$\%$ for testing. For
testing we report the average log-likelihood for each sequence with segment length 0.5s.
For TIMIT we use the predefined test set for testing and split the rest of the data into 95$\%$ for training and 5$\%$ for validation.

During training we use back-propagation through time (BPTT) for 1 second. For the first second we initialize hidden units with zeros and for the subsequent 3 chunks we use the previous hidden states as initialization. 
the temperature $\tau$ starts from a large value 0.1 and gradually anneals to 0.01.

We compare our method with the following methods. For RNN+VRNNs \cite{chung2015recurrent}, VRNN is tested with two different output distributions: a Gaussian distribution (VRNN-GAUSS), and a Gaussian Mixture Model (VRNN-GMM). We also compare to VRNN-I in which the latent variables in VRNN are constrained to be independent across time steps. 
For SRNN \cite{fraccaro2016sequential}, we compare with the smoothing and filtering performance denoted as SRRR (smooth), SRNN (filt) and SRNN (smooth+ $Res_q$) respectively.
The results of VRNN-GMM, VRNN-Gauss and VRNN-I-Gauss are taken from \cite{chung2015recurrent}, and those of SRNN (\upshape smooth+$Res_q$), SRNN (\upshape smooth) and SRNN (\upshape filt) are taken from \cite{fraccaro2016sequential}.
From Table \ref{table_sequential} it can be observed that on both datasets SSNN outperforms the state of the art methods by a large margin, indicating its superior ability in speech modeling.

\subsection{Segmentation and Labeling of Time Series}


To show the advantages of SSNN over HSMM and its variants when learning the segmentation and latent labels from sequences, we take experiments on Human activity  \cite{reyes2016transition}, Drosophila dataset \cite{kain2013leg} and PhysioNet \cite{springer2016logistic} Challenge dataset.Both Human Activity and Drosophila dataset are used for segmentation prediction.

Human activity consists of signals collected from waist-mounted smartphones with accelerometers and gyroscopes. Each volunteer is asked to perform 12 activities. There are 61 recorded sequences, and the maximum time steps $T\approx 3,000$. Each $\m x_t$  is a 6 dimensional vector.

Drosophila dataset records the time series movement of fruit flies' legs. At each time step $t$, $\m x_t$ is a 45-dimension vector, which consists of the raw and some higher order features. the maximum time steps $T\approx 10,000$. In the experiment, we fix the $\tau$ at small value $0.0001$. 

PhysioNet Challenge dataset records observation labeled with one of the four hidden states, i.e., Diastole, S1, Systole and S2. The experiment aims to exam SSNN on learning and predicting the labels. In the experiment, we find that annealing of temperature $\tau$ is important, we start from $\tau = 0.15$ and anneal it gradually to $0.0001$.
\vspace{0.3cm}

\makeatletter\def\@captype{table}\makeatother
\begin{minipage}{.45\textwidth}
	\centering
	\label{table_sequential}
	\begin{small}
		\begin{sc}
			\begin{tabular}{lcccr}
				\hline
				Models                       & \upshape Blizzard      & TIMIT \\
				\hline
				VRNN-GMM & $\geq$ 9107 & $\geq$ 28982 \\
				VRNN-Gauss & $\geq$ 9223 & $\geq$ 28805 \\
				VRNN-I-Gauss & $\geq$ 9223 & $\geq$28805 \\
				SRNN(\upshape smooth+$Res_q$) & $\geq$ 11991 & $\geq$ 60550\\
				SRNN(\upshape smooth) & $\geq$10991 & $\geq$59269 \\
				SRNN(\upshape filt) & $\geq$10846 & 50524 \\
				RNN-GMM & 7413 & 26643 \\
				RNN-Gauss &3539 & -1900 \\
				\upshape Our Method(SSNN) & $\geq$ \textbf{13123} & $\geq$ \textbf{64017} \\ 
				\hline
			\end{tabular}
		\end{sc}
	\end{small}
	\caption{
		\small{
			Average log-likelihood per sequence on the test sets. The higher the better.	}
	}
\end{minipage}
\makeatletter\def\@captype{figure}\makeatother
\begin{minipage}{.55\textwidth}
	\centering
	\vskip 0.05in
	\includegraphics[width=0.89\linewidth]{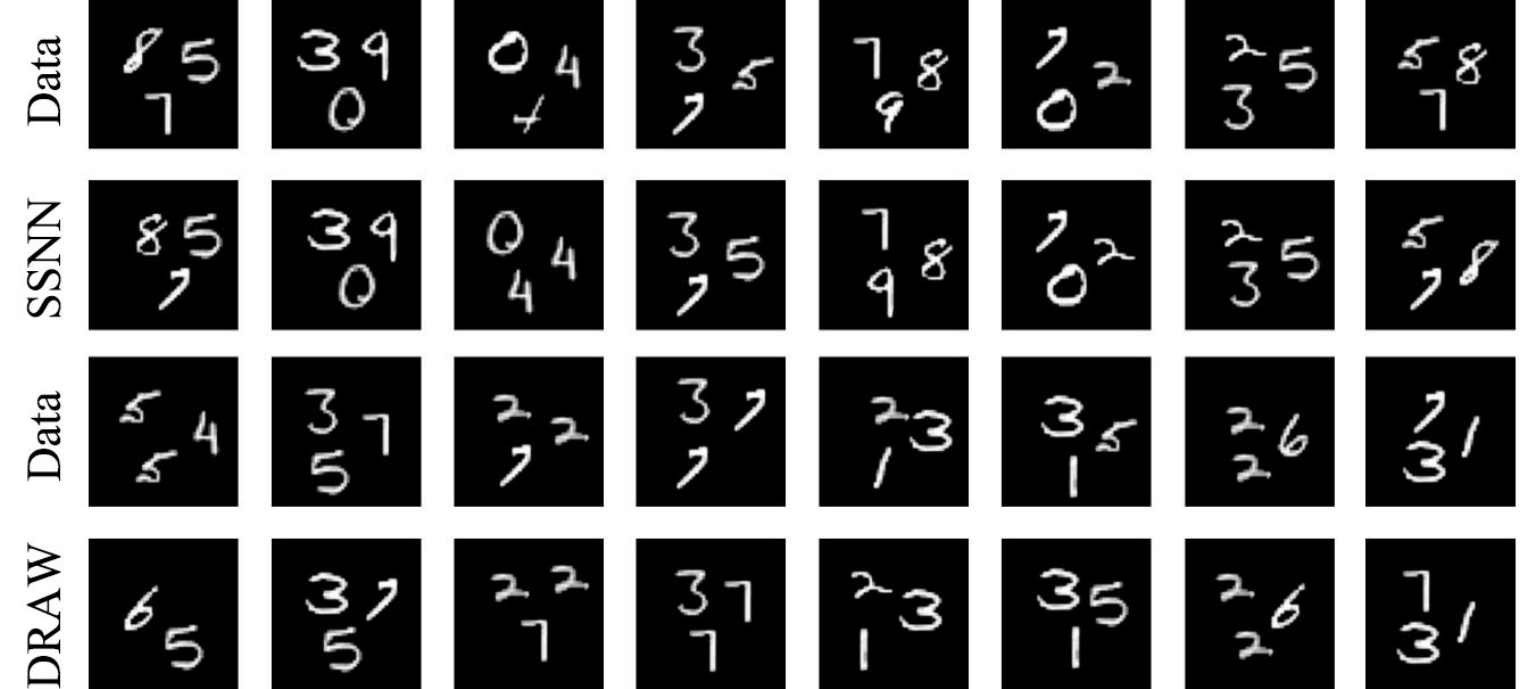}
	\label{fig:non-overlapping}
	\vskip 0.36in

	\caption{
		\small{Visualization and comparison of SSNN and DRAW on multi-object recognition problems. }
	}
\end{minipage}

\begin{table}[htbp]
	\label{table_dro}
	\begin{center}
		\begin{small}
			\begin{sc}
				\begin{tabular}{lcccr}
					\hline
					Models                       &  Drosophila      & Human activity &  Physionet \\ 
					\hline
					HSMM & 47.37 $\pm$ 0.27\% & 41.59 $\pm$ 8.58 \%  & 45.04 $\pm$ 1.87 \% \\
					subHSMM & 39.70 $\pm$ 2.21\% & 22.18 $\pm$ 4.45\% & 43.01 $\pm$ 2.35 \% \\
					HDP-HSMM & 43.59 $\pm$ 1.58\% & 35.46 $\pm$ 6.19\% & 42.58 $\pm$ 1.54 \% \\
					CRF-AE & 57.62 $\pm$ 0.22\% & 49.26 $\pm$ 10.63\% & 45.73 $\pm$ 0.66 \%  \\
					rHSMM-dp & 36.21 $\pm$ 1.37\% & 16.38 $\pm$ 5.06\% & 31.95 $\pm$ 4.12 \% \\
					\upshape SSNN  & \textbf{34.77} $\pm$ 3.70\% &  \textbf{14.70} $\pm$ 5.45\% & \textbf{29.29} $ \pm $ 5.34 \%\\ 
					\hline
				\end{tabular}
			\end{sc}
		\end{small}
	\end{center}
	\begin{small}
		\caption{
			\small{
				Mean and standard deviation of the error rate. For the Drosophila and Human Activity datasets, we report the error rate of segmentation. On the PhysioNet dataset, we report the error rate of latent label prediction.
				The results of subHSMM, HDP-HSMM, CRF-AE and RHSMM-DP are taken from subHSMM \cite{johnson2014stochastic}, HDP-HSMM \cite{johnson2013bayesian}, CRF-AE \cite{ammar2014conditional} and rHSMM-dp \cite{dairecurrent}.
			}
		}
	\end{small}
\end{table}

Specifically, we compare the predicted segments or latent labels with the ground truth,  and report the mean and the standard deviation of the error rate for all methods. We use leave-one-sequence-out protocol to evaluate these methods, i.e., each time one sequence is held out for testing and the left sequences are for training.  We set the truncation of max possible duration $M$ to be 400 for all tasks. We also set the number of hidden states $K$ to be the same as the ground truth.

We report the comparison with subHSMM \cite{johnson2014stochastic}, HDP-HSMM \cite{johnson2013bayesian}, CRF-AE \cite{ammar2014conditional} and rHSMM-dp \cite{dairecurrent}.

For the HDP-HSMM and subHSMM, the observed sequences $\m x_{1:T}$ are generated by standard multivariate Gaussian distributions. The duration variable $d_t$ is from the Poisson distribution. We need to tune the concentration parameters $\alpha$ and $\gamma$. As for the hyper parameters, they can be learned automatically. For subHSMM, we tune the truncation threshold of the infinite HMM in the second level.  For CRF-AE, we extend the original model to learn continuous data. We use mixture of Gaussian for the emission probability. For R-HSMM-dp, it is a version of R-HSMM with the exact MAP estimation via dynamic programming.

Experimental results are shown in Table \ref{table_dro}. It can be observed that SSNN achieves the lowest mean error rate, indicating the effectiveness of combining RNN with HSMM to collectively learn the segmentation and the latent states.

\subsection{Sequential Multi-objects Recognition}
\label{multi-mnist-sec}
In order to further verify the ability of modeling complex spatial dependency, we test SSNN on the multiple objects recognition problem. This problem is interesting but hard, since it requires the model to capture the dependency of pixels in images and recognize the objects in images.
Specifically, we construct a small image dataset including 3000 images, named as multi-MNIST. Each image contains three non-overlapping random MNIST digits with equal probability.

Our goal is to sequentially recognize each digit in the image. In the experiment, we train our model with 2500 images and test on the rest 500 images.First we fix the maximum time steps $T=3$ and feed the same image as input sequentially to SSNN. We interpret the latent variable $d_t$ as intensity and $z_t$ as the location variable in the training images.
Then We train SSNN with random initialized parameters on 60,000 multi-MNIST images from scratch, i.e.,\ without a curriculum or any form of supervision. All experiments were performed with a batch size of 64. The learning rate of model is $1\times 10^{-5}$ and baselines were trained using a higher learning rate $1\times 10^{-3}$. The LSTMs in the inference network had 256 cell units. 

We compare the proposed model to DRAW~\cite{gregor2015draw} and visualize our learned latent representations in Figure~\ref{fig:non-overlapping}. It can be observed that our model identifies the number and locations of digits correctly, while DRAW sometimes misses modes of data. The result shows that our method can accurately capture not only the number of objects but also locations.
\vspace{-2ex}

\section{Conclusion}
\label{sec:conclusion}
In order to learn the structures (e.g., the segmentation and labeling) of high-dimensional time series in a unsupervised way, we have proposed a Stochastic sequential neural network(SSNN) with structured inference.  
For better model interpretation, we further restrict the  label and segmentation duration to be two sequences of discrete variables,  respectively. 
In order to exploit forward and backward temporal information, we carefully design structured inference, and to overcome the difficulties of inferring discrete latent variables in deep neural networks, we resort to the recently proposed Gumbel-Softmax functions. The advantages of the proposed inference method in SSNN have been demonstrated in both synthetic and real-world sequential benchmarks.


\acks{This paper was in part supported by Grants from the Natural Science Foundation of China (No. 61572111), the National High Technology Research and Development Program of China (863 Program) (No. 2015AA015408), a 985 Project of UESTC (No.A1098531023601041), and two Fundamental Research Funds for the Central Universities of China (Nos. ZYGX2016J078 and ZYGX2016Z003). }

\bibliography{ssnn_2017}

\end{document}